\documentclass[sigconf,nonacm]{acmart}
\usepackage{booktabs}
\begin{document}

\title{Generating a Consistent Enterprise with Era by Eon:
Synthesis and Reference-Free Evaluation of Multi-System Business Data}

\author{Benjamin Gruenbaum}
\affiliation{\institution{Eon}\city{}\country{}}
\email{benji@eon.io}
\author{Doron Porat}
\affiliation{\institution{Eon}\city{}\country{}}
\email{doron.porat@eon.io}
\author{Assaf Natanzon}
\affiliation{\institution{Eon}\city{}\country{}}
\email{assaf@eon.io}
\author{Roy Zavida}
\affiliation{\institution{Eon}\city{}\country{}}
\email{roei@eon.io}
\author{Chen Dinachi}
\affiliation{\institution{Eon}\city{}\country{}}
\email{chen.dinachi@eon.io}
\author{Or Itzahary}
\affiliation{\institution{Eon}\city{}\country{}}
\email{or@eon.io}
\author{Omer Niv}
\affiliation{\institution{Eon}\city{}\country{}}
\email{omer.niv@eon.io}

\renewcommand{\shortauthors}{Gruenbaum et al.}

\begin{abstract}
Synthetic relational data is normally produced by a model trained on a
real dataset, and its quality is measured as the distance to that
dataset. This paper describes Era by Eon, a generator that
has no real dataset at either end. Given an industry, a company size, a business model, a set
of business applications, and a random seed, it produces a complete
fictional enterprise: a workforce, a customer base, sales deals,
support tickets, recorded calls, chat messages, and documents, all
consistent with one another. One entity graph is projected into the
native formats of 66 business products, so the same customer appears
in the CRM, the support desk, and the call system under one identity.
Because no real counterpart exists, realism is built in from cited
reference statistics and verified by reference-free measurement: a
five-axis scorecard of 28 statistical checks, an adversarial detector
that hunts for the marks of synthetic generation, and a set of
soundness checks that include a classifier test against an
independently shuffled copy of the data. Because these instruments existed
before the generator was tuned, progress is measured under a fixed
yardstick: over 23 generated companies, mean realism climbed from 60.3
to 99.1, the weakest company from 41.1 to 94.9, and the detector,
which initially flagged 55.2\% of all records, now flags none. The scores hold on a seed never used during
development. A second generator builds relational databases from a
list of business questions. It forces qualifying rows for each
answerable question, adds controlled near misses, and computes exact
labels from the finished tables. The generator runs as a
hosted service at \url{https://console.era.eon.io}. A company built
there to a specification is served through its simulators over MCP
and REST, and the simulators are also published as container images
for offline use.
\end{abstract}

\maketitle

\section{Introduction}

Real enterprise data is confidential. A team building software over a
company's operational systems (its CRM, its support desk, its billing
and communication platforms) cannot take a customer's production
records to test with, and the public substitutes are poor: vendor
sandboxes hold a handful of demonstration records, and manually built
test datasets cover one or two systems and go stale. LLM agents have
sharpened this need, because an agent that reads many business systems
at once has to be exercised on data that behaves like the data it will
find once deployed~\cite{gruenbaum2026era}.

The established route to synthetic data does not apply here. Learned
generators such as the Synthetic Data Vault~\cite{patki2016sdv} and
CTGAN~\cite{xu2019ctgan} train on a real dataset and imitate it, and
quality is then scored by how closely the output matches that dataset
statistically~\cite{hudovernik2024syntherela}. In our setting there is
no dataset at either end. Every generated company is fictional by
design and corresponds to no real one, and training on a customer's
records is exactly what confidentiality forbids. Both
halves of the standard recipe, learning from a reference and
evaluating against a reference, are unavailable.

This paper describes the data-generation system behind Era by
Eon~\cite{gruenbaum2026era}, built for this reference-free setting. A
companion paper describes the agent benchmark that runs on top of the
generated data; the present paper is about the generation itself.
The system described here is publicly available. The hosted
console at \url{https://console.era.eon.io} builds a company to a
specification and serves it over MCP and REST, and each simulator is
published as a container image for offline use
(Section~\ref{sec:availability}). The
system answers three questions.

First, how is a whole enterprise generated so that dozens of systems
agree with one another? The generator produces one object, the entity
graph, that holds everything true about the fictional company. Product
simulators then project the graph: each stores only the portion its
product would keep, expressed in that product's native format, with
record identifiers in the vendor's real syntax. Consistency across systems is not
synchronized; it is a consequence of determinism, because every
simulator rebuilds the same graph from the same seed.

Second, where does realism come from when there is no dataset to
imitate? The generator draws its distributions from a table of
reference targets: named statistics about real business data, drawn
from operational experience and published sources, with every target's
provenance recorded and unsourced targets marked as estimates.
Volumes, values, timestamps, and text are shaped to these targets, and
correlations that real data exhibits, such as larger accounts raising
more tickets, are generated deliberately.

Third, how is the result evaluated without a reference? We use two
instruments that need no paired real data. A realism scorecard tests
the generated records against the reference targets along five axes
and 28 checks. An adversarial detector searches the records for the
known marks of synthetic generation, such as numbered name variants
and timestamps that advance in identical steps. A further set of
soundness checks verifies the data against its own declared design,
including a classifier two-sample test~\cite{lopezpaz2017c2st} against
an independently shuffled copy of the data, which detects missing
inter-column dependencies that marginal statistics cannot
see~\cite{zhang2026dependency}.

The evaluation instruments predate the tuning of the generator, and
every revision pass targeted whatever they reported as worst at that
moment, so development doubles as a controlled experiment. Across 23
generated companies, scored throughout by one fixed scorecard, mean
realism moved from 60.3 to 99.1, the worst company from 41.1 to 94.9,
and the detector's flagged share from 55.2\% to zero. Because all
tuning happened on development seeds, a regression gate rescores every
change on a seed never used during development; the current worst
company on that unseen seed scores 91, with no flagged records.

The paper makes four contributions: (1) the architecture of
Era by Eon, a deterministic generator that produces one
consistent enterprise across
66 product formats from a five-value specification
(Sections~\ref{sec:graph} and~\ref{sec:projection}); (2) a
question-conditioned generator that forces qualifying rows for a
stated list of business questions and recomputes exact labels from the
finished tables
(Section~\ref{sec:synthgen}); (3) a reference-free evaluation
methodology combining realism targets, an adversarial detector, and
design-adherence checks (Section~\ref{sec:eval}); (4) a measured
account of driving a generator from 60.3 to 99.1 under fixed
instruments, including the failure patterns found along the way
(Section~\ref{sec:results}). Section~\ref{sec:availability}
describes how the generated companies can be used, through the hosted
service and through the published simulator images.

\section{Related work}
\label{sec:related}

\textbf{Learned tabular synthesis.} The Synthetic Data
Vault~\cite{patki2016sdv} and CTGAN~\cite{xu2019ctgan} established the
train-and-imitate paradigm for tabular data, and the SDMetrics family
measures quality by comparison with the training data. SyntheRela~\cite{hudovernik2024syntherela} benchmarks relational
generators and finds that the sharpest test of synthetic data is
whether a strong classifier can tell it from the real thing.
Zhang~\cite{zhang2026dependency} exposes a blind spot in the popular
linear classifier two-sample test~\cite{lopezpaz2017c2st}: a table and
a copy of it whose columns were shuffled independently look nearly the
same to it, so broken inter-column dependencies go unnoticed; a
boosted-tree classifier does not share the blind spot.
TabQueryBench~\cite{liu2026tabquerybench} scores synthetic data by how
many analytical queries it answers correctly. All of these
assume a real dataset to learn from or compare against; we adopt their
strongest tests, the boosted-tree classifier and query-level
evaluation, and rework them for a setting with no reference.

\textbf{Specification-driven generators.} TPC benchmark
generators~\cite{tpch} produce data from fixed schemas and analytic
distributions, at any scale and with exact repeatability, but their
uniformity and independence are exactly what makes them unrealistic.
Hollywood~\cite{stoian2026hollywood} builds a synthetic film database
and defends its realism by showing that it misleads a query
optimizer's cardinality estimates~\cite{moerkotte2009qerror} about as
badly as the real database does. Synthea~\cite{walonoski2018synthea}
produces synthetic patient records whose clinical statistics come from
published medical literature, and is the closest precedent for
grounding a generator in cited statistics instead of training data. Our generator extends this
line from one schema to a whole enterprise: many applications, one
shared identity, correlated volumes, a business calendar, documents as
real files, and free text.

\textbf{Benchmark environments.} Enterprise agent benchmarks such as
$\tau$-bench~\cite{yao2024tau}, CRMArena~\cite{huang2025crmarena},
TheAgentCompany~\cite{xu2024theagentcompany}, and
WorkArena~\cite{drouin2024workarena} hand-build their environments,
which bounds their size and their realism to what their authors could
author. Text-to-SQL benchmarks showed that with real data, much of
the difficulty comes from the data rather than the query language:
BIRD~\cite{li2023bird} includes
dirty values and unanswerable questions, and
Spider~2.0~\cite{lei2025spider2} moved evaluation to enterprise
warehouses. We inject controlled doses of dirty values and generate
labeled unanswerable questions at generation time.

\section{One specification, one graph}
\label{sec:graph}

\begin{figure*}
\centering
\includegraphics[width=\textwidth]{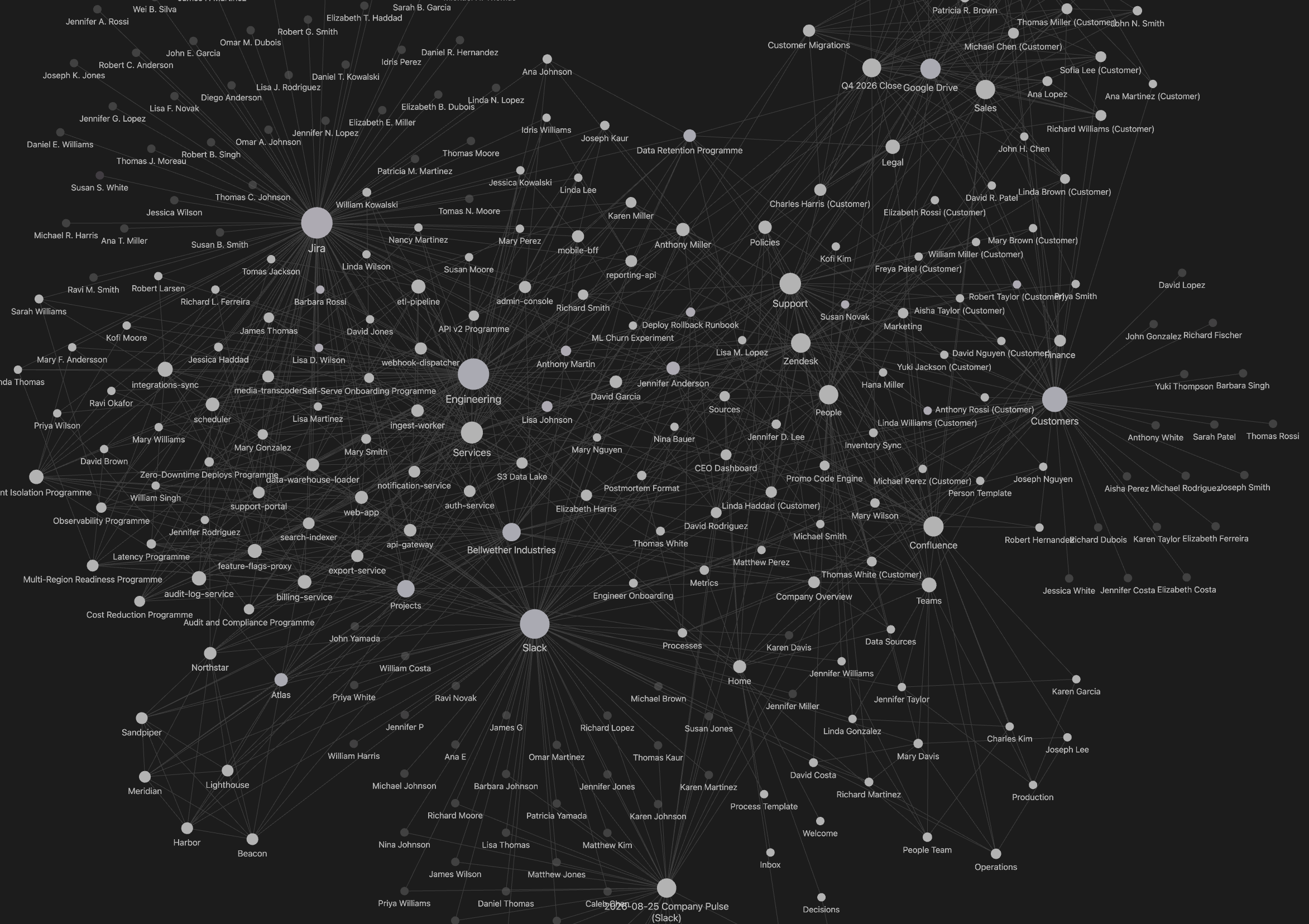}
\caption{The entity graph of one generated company, Bellwether
Industries, drawn from records read back through the simulators' MCP
tools. Each node is one entity: an employee, a customer, a team, a
project, a service, or a business system such as the issue tracker or
the chat workspace. Each edge is a reference between two records.
Every system's records connect back to the same people and accounts,
because all systems are projections of this one graph.}
\label{fig:graph}
\end{figure*}

\subsection{The specification and the entity graph}

The input to the generator is a five-value specification: an industry
(one of ten, from e-commerce to manufacturing), a company size (one of
six tiers, from 10 to 20{,}000 employees), a business model (B2B, B2C,
or B2B2C), the set of business applications the company runs, and a
random seed. From this specification the generator produces the entity
graph: a single in-memory object that holds everything true about the
fictional company. Its identity, its people, its customers, and every
record of business activity live in this one graph. Every downstream
artifact is a projection of it. Figure~\ref{fig:graph} shows the graph
of one generated company, drawn from records read back through the
running simulators.

Generation proceeds in dependency order, so that later records can
reference earlier ones. The workforce comes first. An employee record
carries a job-ladder position and a manager, a hire date sampled from
realistic tenure curves, an employment type, and a home country that
brings its own holidays and working hours; the department mix follows
the company's size and growth stage. A fraction of the workforce has
already departed, and the records those people created stay behind
under an owner who no longer exists. The customer base comes next,
assigned to sales representatives as territories, and business
activity grows around it: contacts, then deals carrying their complete
stage history, then documents, chat traffic, support tickets, recorded
calls, and campaigns. A ticket's requester is therefore always a contact of a
real account, and the owner of a deal is always a member of the sales
team.

The graph also carries the texture of a company that has actually been
operating. The CRM contains typographical duplicates. A long list of
prospect accounts that never became customers sits beside the real
customer book. Service accounts that belong to software rather than
people appear among the users. Knowledge-base articles are written
from the company's own tickets, and the company runs infrastructure of
its own, with services that deploy and occasionally fail. These
imperfections are generated deliberately, at a controlled rate,
because software that runs on enterprise data must survive them.

\subsection{Determinism by construction}
\label{sec:determinism}

The same specification and seed produce the identical company, byte
for byte, on any machine. Everything else in the system leans on this
property. It is how 66 independent simulators serve one consistent
company without ever exchanging data, and it is how a generated corpus
can be committed to a repository and checked in continuous integration
by regenerating it and comparing bytes.

Plain determinism is easy. The hard part is keeping the output stable
while the generator's code keeps changing. A pseudo-random generator
hands out numbers in a fixed order, so every value depends on how many
values were drawn before it. If the whole generator shared one stream
of numbers, any code change that adds a single extra draw would shift
every draw after it. Add one attribute to employees, and every
customer name, ticket timestamp, and call transcript drawn later would
change. The corpus diff would be enormous and would say nothing about
what the change actually did.

Two mechanisms keep changes contained.

The first works at the level of subsystems. The generator is organized
into subsystems, each building one aspect of the company: who the
customers are, which of them left, what the chat rooms say, what sits
in the file drive. A subsystem is an aspect of the graph, not a
product. There is no call-recorder generator and no CRM generator. A
recorded call is generated once, in the graph, and the call system and
the CRM each project that same call (Section~\ref{sec:projection}).

Each subsystem gets its own random generator, called its stream,
freshly seeded at the subsystem's start with the company seed combined
with a fixed constant for that subsystem
($\mathit{seed} \oplus o_s$). One number still determines the whole
company, but no subsystem continues another's sequence, so no
subsystem's values depend on how many draws another made. There are 21
subsystems, each with its own stream: identity (names, addresses,
domains), structure (how many of everything), distribution (amounts
and durations), temporal (timestamps), lifecycle (stage histories),
correlation (cross-field coupling), content (text), geography,
customers, retention, journey, attributes, time-off, segment,
prospecting, hygiene, imperfection, chat, desk, support, and drive.

The isolation applies to randomness, not to data. Generation runs in
dependency order, and each subsystem reads what earlier subsystems
produced. Customers illustrate this. The generated company sells to
other companies. Each customer is stored in the graph as a single
record: a company name, a web domain, an industry, and the contact
people who work there. Deals, tickets, calls, and files all refer to
these records. When the call subsystem runs, the customer records
already exist. It reads them as input and uses randomness only for its
own decisions: which calls take place, at what times, between which
participants, and with what content. A call refers to a customer
record; it never draws that record's contents anew. A change to the
customer names therefore appears in every call that mentions a
customer, because calls are built from the customer records. The
reverse cannot happen: no change to call generation can affect which
customers exist or how they are named, because those values come from
random draws that the call subsystem never touches.

Stream boundaries are placed where one aspect must hold still while
another varies. Who the customers are has its own stream, so resizing
any simulator's data cannot change the set of customer records. Which
customers have left the company has its own stream, so the CRM, the
support desk, and the file store all name the same departed customers.
Prospects have their own stream because of their number. A prospect is
a potential customer that never bought anything; sales teams import
them as purchased lists, and they make up most of the accounts in a
real CRM. The generator creates about ten of them per customer, which
costs more random draws than anything else in the graph, so on a
shared stream any change to their number would shift everything drawn
after them.

The second mechanism works at the level of single values, inside a
subsystem. A fine-grained value is not drawn from a running sequence
at all. It gets its own tiny generator, seeded by a key that says what
the value is for: the purpose, the company seed, and the record it
belongs to. One sales representative's call volume, for example, comes
from the key \texttt{call-load/\textit{seed}/\textit{rep}}. Such a
value depends only on the seed and on which record it belongs to. What
was generated before it, and in what order, does not matter. New
record types and new attributes can therefore be added without moving
a single existing value.

Together the two mechanisms bound what a change can touch. Data still
flows downstream: change how customers are generated, and every record
that quotes a customer changes with it, a call transcript that names
one, a ticket filed by one of its contacts. What cannot happen is
unrelated churn, because no other subsystem's random draws have moved.
The same calls still happen at the same times between the same
participants; only the quoted customer details differ. The
regeneration diff therefore contains the changed records and the
records that reference them, and nothing else.

Time is pinned the same way. No timestamp is read from the wall clock.
Every date is measured backward from one configured anchor, the
company's ``today'', and holidays and quarter ends are computed by
rule from the anchor's year. With the anchor pinned to a fixed date,
regenerating a year later reproduces the identical company; this is
how the committed corpus and the test suites run. With the anchor set
to track the calendar day, a live deployment stays current: the whole
history slides forward one day per day and stays internally
consistent, because every date keeps its distance from the anchor.

\subsection{Realism from reference targets}
\label{sec:targets}

The generator is not trained on any dataset. Instead, it reads its
numbers from a table of reference targets. A reference target is one
named statistic about real business data, for example how ticket
priorities are distributed, or how deal sizes are distributed. The
table records where each number comes from. Seven targets come from
published industry statistics and from operational experience. Three
describe only the shape a distribution should have, such as a few
accounts holding most of the deals. Three could not be sourced and are
marked as estimates (Table~\ref{tab:targets}).

The same table drives both generation and evaluation. The generator
draws its distributions from the targets, and the scorecard of
Section~\ref{sec:eval} measures the finished data against the same
targets, read from the same definition. The generator therefore cannot
drift away from what the scorecard checks.

Real business data also contains flaws: duplicated records, empty
fields, inconsistent spellings. The generator plants such flaws on
purpose, at a fixed rate of 3\% of records, so the data carries the
flaws that software running on it must survive.

\begin{table}
\caption{The reference targets. Each target is a named statistic with
recorded provenance; ``estimate'' marks targets for which no source
was available.}
\label{tab:targets}
\begin{tabular}{lll}
\toprule
Target & Kind & Provenance \\
\midrule
Ticket priority mix & categorical & industry \\
Ticket status mix & categorical & industry \\
Ticket channel mix & categorical & industry \\
Deal stage outcome mix & categorical & industry \\
Lead source mix & categorical & estimate \\
Call sentiment mix & categorical & estimate \\
Deal amount distribution & shape & industry \\
Account employee counts & shape & industry \\
Call duration distribution & shape & estimate \\
Deals per account & count & structural \\
Tickets per account & count & structural \\
Contacts per account & count & structural \\
Timestamp calendar structure & temporal & industry \\
\bottomrule
\end{tabular}
\end{table}

The reference targets determine the statistical character of each
generated company. Ticket priorities follow observed support patterns,
with roughly half of the tickets marked as normal and 8\% as urgent.
Deal amounts follow a long-tailed log-normal distribution whose scale
changes with company size. For example, a mid-market B2B software
company has a median deal amount of about \$48{,}000, while smaller and
larger companies use lower and higher values. Activity is also
unevenly distributed. A small number of customers account for much of
the sales and support activity, while many customers have only
occasional records.

Realism also depends on relationships between fields. Larger customers
tend to open more support tickets. Customers with unresolved urgent
tickets have a higher risk of not renewing. The sentiment of a sales
call is related to the eventual outcome of the deal. The generator
creates these relationships explicitly, and Section~\ref{sec:eval}
checks whether they appear in the finished data.

Generated events follow the working patterns of the people involved.
In the current reference model, 78\% of events occur during local
business hours and 6\% occur on weekends. Activity increases near the
end of each quarter. Events also respect causal order, so a call about
a deal cannot occur before the deal has opened.

The generator assigns a single topic to each business event and uses
it throughout the related text. A ticket subject agrees with its body,
and a call transcript remains consistent with the corresponding deal
notes. Templates combine common business language with vocabulary
from the selected industry. Across the corpus, vocabulary grows with
the amount of generated text instead of repeating a small fixed set of
phrases.

The generation process remains practical at large scales. A company
with 20{,}000 employees and approximately 250{,}000 records takes about
nine minutes to generate and occupies about 25\,MB on disk.

\section{Projection into 66 product formats}
\label{sec:projection}

\subsection{Simulators as projections}

Once the entity graph has been generated, each simulator converts the
relevant parts of the graph into the records used by one business
product. A simulator is a service that reproduces that product's
interface and data model. It answers the vendor's own REST
API, with the vendor's query language and authentication header, and
exposes the same operations as an MCP server, so an agent written
against the real product can be pointed at the simulator instead. The
system currently includes 66 simulators,
covering products such as Salesforce, Zendesk, HubSpot, Slack, Jira,
Gong, Stripe, Google Drive, and Amazon S3. The simulators do not
exchange records with one another. Each independently rebuilds the
company graph from the specification and seed, or loads a previously
generated copy, then selects the records its product would store. This
gives every product a different view of the company while keeping
shared entities and events consistent across products.

Each simulator exposes only the information that its product would
normally contain. For example, Gong stores a recorded sales call with
its participants and full transcript. Salesforce stores the same call
as an activity linked to the corresponding customer, contact, and
deal. Both records come from the same call in the entity graph, so
they agree on the participants, time, and customer even though their
fields differ. Product-specific records remain in the appropriate
systems. A duplicate customer created by a CRM data-entry error appears
in the CRM but not in Gong, while potential customers appear only in
products that maintain customer lists.

The company specification includes a portfolio, which lists the
business applications the company uses. The portfolio does not change
the entity graph; the generator always builds the complete company.
Instead, it controls what each simulator exposes. If the company does
not use a particular product, that product's simulator returns a valid
but empty workspace. Records created by an integration appear only
when both products are present. For example, Salesforce includes
activities created by the Gong integration only when the portfolio
includes Gong. Without Gong, the calls still exist in the entity
graph, but Salesforce contains no Gong-specific records. Companies
generated with the same industry, size, business model, and seed
therefore retain the same people, customers, deals, calls, and tickets
even when their application portfolios differ.

\subsection{Shared and vendor-shaped identifiers}

The entity graph assigns one internal identifier to each person,
customer, and business event. Most simulators expose these shared
identifiers by default. The same person or event can therefore be
recognized directly across products. Simulators also publish a stable,
opaque identity key in fields intended for external identities. That
key provides a second way to reconcile records without relying on
names or email addresses.

Some vendor clients enforce their own identifier syntax. For these
clients, a simulator can replace the shared identifier with a
vendor-shaped identifier. This mode is optional and has not been
adopted by every simulator. When it is enabled, the identifier is a
deterministic function of the company, graph identifier, and vendor
format. The simulator checks the complete identifier set for
collisions, resolves any collision by deterministic re-probing, and
keeps a reverse mapping to the graph identifier.

Cross-product joins therefore use the identity information that each
product exposes: shared identifiers in the default mode, the stable
external identity key, or an email address when email is the product's
native identity. Vendor-shaped identifiers remain specific to the
product that issued them.

\subsection{Documents as real files}

The generator creates recognizable business documents rather than
generic text files. Each document follows a template for a specific
purpose. Examples include security reviews, renewal proposals, pricing
worksheets, data-processing agreements, migration runbooks, employee
records, and support summaries. The template determines the expected
sections and structure. A security review contains findings,
severities, owners, and a decision. A migration runbook contains
prerequisites, execution steps, rollback instructions, and
verification checks. A pricing worksheet contains line items whose
values add up to the associated deal amount.

Document content is tied to the entity graph. The author is an
employee from the generated company. Customer names, contacts,
domains, deal values, support tiers, and dates come from the records
associated with the file. A renewal proposal therefore names the same
customer and deal that appear in the CRM. Its commercial values agree
with that deal. The corresponding pricing worksheet uses those values
in its calculations. Local details that do not appear elsewhere in
the graph are generated from the file identifier. This keeps shared
facts consistent while allowing documents about different customers
to contain different findings, risks, clauses, and action items.

The templates contain several alternative passages and data patterns.
The file identifier selects among them and controls their order. This
produces variation without breaking reproducibility. Some document
types also contain fictional sensitive values in realistic formats,
such as payment-card numbers, national identifiers, bank accounts,
and access keys. These values use documented test ranges or markers,
so they resemble the data found in enterprise files without belonging
to a real person or account.

The renderer converts this content into valid DOCX, XLSX, and PDF
files. Tables remain tables in spreadsheets, while headings and
paragraphs retain document structure in Word files. Fixed archive
metadata and timestamps make the resulting bytes deterministic. The
same file can therefore appear unchanged in Amazon S3, Google Drive,
Box, Dropbox, and SharePoint. Text and Office documents receive the
full business-content model. Images, audio, and video are currently
valid deterministic files with simpler generated content.

\section{Question-conditioned database generation}
\label{sec:synthgen}

Alongside the 66 product simulators, Era by Eon includes a
second generator for tables used by a company's own applications and
analytics systems. These additional tables have schemas defined by
the company's information needs rather than by a software vendor.
They are generated separately from the product projections and may
form one database or several related databases.

Business questions are supplied as input rather than
generated from an existing schema. The person configuring a dataset
writes them in a text file, with one question per line. The questions
may ask for filtered records, aggregates, temporal comparisons, or the
absence of a related record. Examples include ``Find all contracts
that expire in the next 90 days,'' ``What is the total contract value
for each customer?'' and ``Which customers have unpaid invoices older
than 30 days?''

The language model reads the complete question set and
infers the entities and relationships needed to answer it. These
examples require related tables for customers, contracts, and
invoices. Contract expiration dates support the first question.
Contract values and customer relationships support the second.
Invoice dates and payment status support the third. The resulting
output contains the database schemas, populated tables, and exact
answers computed from those tables. It also identifies questions that
the generated databases cannot answer and records the missing table
or column.

\textbf{Designing the database.} A language model is used only during
the design phase. It reads the questions, identifies the business
domain, and proposes the databases and entities needed to answer them.
It then produces a concrete schema. Each column has a type and a
generation plan that describes its domain, distribution, null rate,
and relationships. Each question is linked to the tables, joins,
filters, and aggregation needed to answer it.

The proposed design passes through deterministic validation and
repair. The validators check keys, references, table dependencies,
column domains, and the instructions associated with each question.
Mechanical mistakes are repaired in code. A design that remains
invalid or produces poor data is returned to the language model with
specific feedback. The number of redesign attempts is bounded, and
the best valid design is retained. Once this phase is complete, the
language model is no longer used.

The schema is created once for each question list. It is not
created for every fictional customer, and the random seed is not an
instruction to the language model. During the design run, however,
the pipeline populates each candidate schema using the selected seed
and measures the quality of the resulting data. This score can affect
which candidate schema is retained.

The language-model phase is not deterministic. Running the
complete pipeline again may produce a different schema, even with the
same questions and seed. The accepted application description and
data model are therefore saved with the generated dataset. Exact
regeneration starts from this stored data model. Given the stored
model and the same seed, the seeded generator reproduces the same rows
and labels. A different seed changes the rows and labels while leaving
the stored schema unchanged.

The design may also contain questions that the database cannot answer.
Each such question must name the missing table or column. The
validator accepts the question only when that information is absent
from the final schema. This makes the absence verifiable from the
database design itself.

\textbf{Generating the rows.} Seeded code fills the tables in
dependency order, creating referenced rows before the rows that point
to them. It then forces a chosen number of rows to satisfy the
conditions of each question. This operation, called planting, ensures
that an answer does not become empty merely by chance.

For questions with at least two conditions on the same table, the
generator also creates near-miss rows. Each row satisfies all but one
of those conditions. Consider a question about failed jobs in Europe
that have not been deleted. One near miss may describe a failed
European job that was deleted, while another may describe an
undeleted failed job in a different region. These rows ensure that
each tested condition changes the result.

Controlled data flaws are added after the tables have been populated.
They include inconsistent capitalization, extra whitespace, and
placeholder text. Key columns, relationship columns, and columns used
to compute an answer are protected from these changes. The result
contains realistic imperfections without breaking the relationships
or labels that define the dataset.

\textbf{Computing the labels.} The generator executes every question
against the completed tables. This final query result becomes the
stored label. It includes both the rows created for the question and
any other rows that happen to satisfy the same conditions. Each label
also records how it should be compared. Lists are treated as sets
unless their order matters, mappings are compared by key, and
fractional values include a numeric tolerance.

\textbf{Generating successive days.} A day state is one snapshot of a
generated database. Each new day starts from the previous snapshot and
applies a deterministic set of changes. By default, the generator
inserts rows equal to 5\% of each table, updates another 5\%, and
deletes 2\%. Updates change one mutable field and leave keys and
references intact. A deletion uses a soft-delete field when the schema
provides one; otherwise, a row is removed only when no other row
references it. The generator records which rows were inserted,
changed, or removed and recomputes all labels for the new snapshot.

\section{Evaluation without a reference}
\label{sec:eval}

Most synthetic-data evaluation compares generated records with the
real dataset used to train the generator
~\cite{patki2016sdv,hudovernik2024syntherela}. Era by Eon has no such
dataset. Its checks use three sources of evidence instead: the
generation plan, transformations of the generated rows, and the
reference targets in Section~\ref{sec:targets}. The checks in
Section~\ref{sec:soundness} apply to the question-conditioned
databases. The scorecard and detector in the following sections apply
to the enterprise entity graph.

\subsection{Checks against the generation plan}
\label{sec:soundness}

\textbf{Plan adherence.} Every column plan states what the finished
column should contain. It defines the allowed categories or numeric
range, the expected share of missing values, and any intended
distribution or relationship. After generation, the rows are checked
against those declarations. The check reports values outside their
domain, unexpected missing values, absent categories, incorrect
distribution shapes, and relationships whose number of child records
falls outside the planned range.

\textbf{Question specificity.} Every condition in a question should
affect its answer. The system tests this by executing modified versions
of the question against the finished data. These versions remove one
condition, omit a join, ignore deletion status, or aggregate all rows
without filtering. If a modified question returns the same answer,
the generated rows do not make the omitted condition meaningful. The
question is then reported for redesign or new planting.

\textbf{Label integrity.} Each answerable question has a stored label:
the answer recomputed from the completed tables. Mechanical checks
verify that this answer is unique and can be compared fairly. For a
question asking for the ten largest contracts, the tenth and eleventh
contracts must not have the same value. Otherwise, two different
lists could both be correct. Fractional answers, such as averages,
must include a numerical tolerance to allow for rounding. The checks
also flag questions that contain their own answer and pairs of
questions that produce the same nontrivial answer.

\textbf{Dependency structure.} Matching each column separately does
not ensure that values are combined realistically within a row. A
generated table might contain the intended mix of small and large
customers and the intended distribution of deal values, yet assign
deal values independently of customer size. Both columns would look
correct on their own, but the expected relationship between them
would be missing. The system tests for this problem by independently
shuffling every column. Shuffling preserves each column's values and
frequencies while removing relationships between columns. A
boosted-tree classifier then tries to distinguish the original rows
from the shuffled rows. If it cannot, the intended relationships are
not visible in the generated data
~\cite{lopezpaz2017c2st,zhang2026dependency}.

\textbf{Cardinality-estimation error.} We use cardinality estimation
as a second measure of structure in the generated tables. For each
query, we first compute the exact number of matching rows. We then
estimate the same number under two simplifying assumptions: values are
uniformly distributed, and conditions on different columns are
independent. We report the difference as q-error
~\cite{moerkotte2009qerror}, defined as the ratio between the larger
count and the smaller count. A q-error of one indicates an exact
estimate, while a value of four indicates a fourfold error. Large
errors arise when values are unevenly distributed or when columns are
related. The test therefore determines whether the generated tables
contain statistical structure that an independent, uniform model
cannot capture. It requires no real reference dataset because both the
estimate and the exact count are computed from the generated tables
~\cite{stoian2026hollywood}.

\subsection{The realism scorecard}
\label{sec:scorecard}

The scorecard turns the properties described in
Section~\ref{sec:targets} into measurable results. It runs after the
entity graph has been completed. Its 28 checks are divided into five
groups. Each check compares the generated data with a reference value,
an acceptable range, or a consistency rule.

\textbf{F1: individual fields.} These checks examine the values in one
field. For a categorical field, such as ticket priority, the scorecard
calculates the percentage of records in each category and compares
those percentages with the reference distribution. For a numerical
field, such as deal value, it measures the spread of the values. This
reveals fields whose average is plausible but whose values are too
similar. The group contains eight checks covering tickets, deals,
calls, and customer-company sizes.

\textbf{F2: relationships between fields.} These checks measure
whether two related fields change together. For customer size and deal
value, the scorecard ranks customers and deals from smallest to
largest and compares the two rankings. It uses the same method for
urgent tickets and renewal risk. For call sentiment, it compares the
percentage of positive calls associated with successful deals with the
percentage associated with other deals. Additional checks examine
negative customer ratings and the relationship between employee count
and revenue. Each relationship must be visible without becoming an
exact formula.

\textbf{F3: time.} These checks begin by converting every event to the
local time of its customer. The scorecard then measures the percentage
of events that occur during business hours and on weekends. It also
counts repeated timestamps and measures whether the intervals between
events vary. The final check examines related events in pairs and
calculates the percentage that appear in the correct order. A ticket
response, for example, must occur after the ticket was created.

\textbf{F4: distribution across customers.} The scorecard counts the
deals, tickets, and contacts associated with every customer. It sorts
the customers by each count and calculates the share held by the
busiest 10\%. That share is compared with the reference value. Deals,
tickets, and contacts are measured separately, producing three checks.

\textbf{F5: text.} These checks count distinct ticket subjects,
messages, and transcript lines. They measure how many tickets contain
a description and how many contain comments. Vocabulary size is
evaluated relative to the total amount of text, so larger datasets are
expected to contain more distinct words. The final check reads each
ticket as one record and verifies that its subject, description, and
comments refer to the same issue.

Each check requires enough records to support its calculation. A check
with too little data is marked as unmeasured and is excluded from the
score. The report also states how many checks ran. This prevents a
score based on a small subset of the data from appearing complete.

The development process uses fixed seeds for tuning, which creates a
risk of overfitting to those companies. Continuous integration
therefore scores every company on a separate seed that was not used
during development and rejects changes that fall below the required
score.

\subsection{The adversarial detector}
\label{sec:detector}

The scorecard measures properties for which a reference value or
consistency rule has been defined. The adversarial detector performs a
different test. It searches the completed entity graph for simple
patterns that reveal how the data was generated.

For example, generated names may repeat with a different number
appended to each copy. Categories may appear in almost identical
proportions because values were assigned in a fixed cycle. Timestamps
may all be equal or separated by perfectly regular intervals. Every
customer may also receive exactly the same number of contacts, deals,
or tickets.

Other patterns appear in text and numerical fields. The detector finds
sentences repeated across many records, tickets with no description,
and fields that contain the same value throughout the dataset. It also
finds deal values confined to an unusually narrow range.

For every detected pattern, the report identifies the affected records
and provides representative examples. It also reports the percentage
of records affected by at least one pattern. A value of zero means
that none of the patterns covered by the detector was found. It does
not prove that the generated data is indistinguishable from real
enterprise data.

\section{Results}
\label{sec:results}

\subsection{Measured improvement under fixed instruments}

We used a paired experiment to measure the effect of the realism
mechanisms. We selected 23 fixed company specifications covering ten
industries and six company sizes. Each specification was generated
twice with the same random seed. The realism profile was the only
setting that changed between the two runs.

The first run used the legacy profile. This compatibility mode
reproduces the generator's behavior before the realism mechanisms were
added. It uses uniform or fixed record counts, independent field
values, simple timestamp patterns, and a small collection of repeated
text templates. The second run enabled the complete realism profile.
It uses the reference distributions, uneven record counts, local
business calendars, causal event sequences, relationships between
fields, varied text, plausible identities, and cross-system customer
histories.

We evaluated both versions with the same final scorecard and
adversarial detector. Because each pair used the same company
specification and seed, the comparison isolates the effect of the
realism profile from differences in industry, company size, or random
input.

\begin{table}
\caption{Realism scores of 23 generated companies under one fixed
scorecard, before and after the generator revisions. Overall values
include all 23 companies. Axis means include only companies with
enough data to score that axis: F1 uses 5 companies in each version,
F2 uses 5 original and 7 revised companies, F3 and F5 use all 23, and
F4 uses 4. Each axis scores 0--100.}
\label{tab:axes}
\begin{tabular}{lrr}
\toprule
 & Original & Revised \\
\midrule
F1 marginal & 63.6 & 97.7 \\
F2 joint & 54.9 & 97.1 \\
F3 temporal & 83.2 & 99.7 \\
F4 structural & 36.6 & 93.9 \\
F5 content & 34.3 & 98.6 \\
\midrule
Overall (mean) & 60.3 & 99.1 \\
Overall (worst company) & 41.1 & 94.9 \\
Detector-flagged records & 55.2\% & 0.0\% \\
\bottomrule
\end{tabular}
\end{table}

The mean overall score rose from 60.3 to 99.1, and the lowest company
score rose from 41.1 to 94.9. The largest gains appeared in structural
and content quality. Under the legacy generator, customers had similar
numbers of related records and text repeated across the corpus. The
revised generator produced concentrated relationship counts and much
greater text variety.

The detector found 137 artifacts in the legacy companies. These
included 40 cases of identical relationship counts, 33 cases of
evenly spaced timestamps, 23 cases of missing text, 21 cases of
constant timestamps, and 13 fields with nearly uniform category
frequencies. In total, 55.2\% of legacy records were affected by at
least one artifact. The detector found none in the revised companies.
On the separate seed used by the continuous-integration gate, the
lowest company score was 91 and no records were flagged.

\subsection{What the instruments caught}

\textbf{Transcript repetition.} One hyperscale company contained
74{,}167 transcript turns generated from only 42 fixed sentences. The
most common sentence appeared 2{,}936 times, and a single call could
combine unrelated subjects. The revised generator assigns one subject
to each call and renders every turn from templates tied to that
subject. The same company now contains 19{,}794 distinct utterances,
and the most common sentence appears 631 times.

\textbf{Small samples.} Early versions of some scorecard checks
measured sampling noise rather than generator quality. In one
12-customer company, the same correlation was 0.55 under one seed and
$-0.26$ under another. The final scorecard uses rank correlations,
minimum sample sizes, and wider tolerances for small samples. All
legacy and revised results in Table~\ref{tab:axes} were recomputed
with this final version.

\textbf{Why both instruments are needed.} The scorecard and detector
examine the data in different ways. The scorecard summarizes
statistical properties as numerical scores. The detector identifies
specific records that contain known signs of automatic generation.

A company can receive strong statistical scores while still
containing an obvious artifact. For example, adding sequential numbers
to employee names does not change deal distributions, customer
activity, or event timing, but it makes the generated origin easy to
recognize. The detector reports those names directly.

The reverse can also occur. Customer activity may vary enough to avoid
the detector's rule for identical record counts, yet remain much more
evenly distributed than the reference target. The structural score
identifies this broader statistical problem.

Some failures, including repeated text and regularly spaced
timestamps, are examined by both instruments. The scorecard measures
how widespread the problem is, while the detector identifies the
affected records. We therefore report both results. A high score
cannot hide an obvious generation artifact, and an empty detector
report cannot hide unrealistic statistical structure.

\subsection{Corpus regeneration}

The committed corpus contains 83 generated companies and occupies
2.0\,GB. Sixty companies cover every combination of the ten industries
and six size tiers, and the remaining 23 are individually specified
companies. Continuous integration scores the generator on an unseen
seed and fails if any company falls below the required fidelity score.
A separate job regenerates the committed corpus and reports byte-level
drift. This separates unintended drift from deliberate changes that
require a new corpus release.

\section{Availability}
\label{sec:availability}

The generator and the simulators are available as a hosted service,
Era by Eon, at \url{https://console.era.eon.io}~\cite{eraconsole}.
A user describes a company with the specification of
Section~\ref{sec:graph}: an industry, a company size, and the set of
business systems the company runs. The service generates the company
and starts one simulator per selected system. Each simulator answers
the vendor's own REST API and the same operations over MCP, and the
service issues one token per company, or one token scoped to a single
system, with which an agent authenticates against them. A company can
be built from the browser, from the \texttt{era} command-line tool, or
through the console's HTTP API and its own MCP server. Access is
granted per account and is free for builders.

Every environment is synthetic, and no customer data is involved. The
data plane is read-only. An agent reads the company but cannot write
into it, so the company does not change under an agent that works in
it. A hosted company is exposed at fixed day states: the company at
inception, with empty systems; the company today, with its full
history up to the anchor date; and the company one quarter later.
Moving between day states regenerates the company from its
specification, as described in Section~\ref{sec:determinism}.

For continuous integration, each simulator is also published as a
container image on Docker Hub, under the \texttt{erabyeon} namespace,
for example \texttt{erabyeon/slack}. An image runs offline and answers
MCP and REST on a local port. It builds its company from the seed and
anchor date set in its environment, so several images composed
together serve one consistent company without exchanging data, as
Section~\ref{sec:projection} describes. Pinning the seed, the anchor
date, and the image tag keeps the served company unchanged from one
run to the next.

Because every environment is synthetic, the service is not a
substitute for a staging system that carries real data. It is a place
to build, test, and demonstrate software that reads enterprise systems
before that software meets a customer's records.

\section{Limitations}
\label{sec:limitations}

Era by Eon can only be as accurate
as its reference targets. Ten of
the thirteen targets have recorded provenance from industry statistics
or structural knowledge, while three are marked as estimates. An
incorrect target will produce an incorrect distribution, and a
scorecard that reads the same target will not reveal the problem.

Generated text is based on seeded templates. The templates produce
varied and internally consistent text, but they cannot reproduce the
full range of style, ambiguity, error, and context found in human
writing. A stronger detector may therefore distinguish the generated
text from real business communication.

The reported evaluation compares two versions of the same generator.
It does not establish how Era by Eon compares with other
specification-driven generators. The detector has also not been
calibrated against a large corpus of real enterprise records, because
such records are the confidential data the system is designed to
avoid.

\section{Conclusion}

Era by Eon generates a complete fictional company from a compact
specification and a seed. A shared entity graph keeps people,
customers, and business events consistent across 66 product
simulators. Reference targets shape the distributions, relationships,
timing, and text without training on confidential customer data.

The question-conditioned generator extends the same approach to
company-specific databases. It turns business questions into a
validated schema, deterministic generation plan, populated tables, and
labels computed from the final rows. Successive day states add
controlled change while preserving keys and relationships.

The evaluation uses the generation plan, transformed copies of the
generated rows, and recorded reference targets in place of a paired
real dataset. The measured comparison shows that these checks exposed
problems in structure, timing, and text that simple field-level
statistics missed. The result is reproducible multi-system enterprise
data generated without access to customer records, with its
statistical assumptions stated explicitly. The generated
companies are available to others through the hosted service at
\url{https://console.era.eon.io} and through per-system container
images for offline use.

\bibliographystyle{ACM-Reference-Format}
\bibliography{era_generation}

\end{document}